\documentclass[fleqn,10pt]{wlscirep}
\usepackage[utf8]{inputenc}
\usepackage[T1]{fontenc}
\usepackage{float}
\title{Analysis of Line Break Prediction Models\\ for Detecting Defensive Breakthrough in Football}

\author[1]{Shoma Yagi}
\author[2]{Jun Ichikawa}
\author[1,*]{Genki Ichinose}
\affil[1]{Department of Mathematical and Systems Engineering, Shizuoka University, Hamamatsu, 432-8561, Japan}
\affil[2]{Faculty of Informatics, Shizuoka University, Hamamatsu, 432-8011, Japan}

\affil[*]{ichinose.genki@shizuoka.ac.jp}

\begin{abstract}
In football, attacking teams attempt to break through the opponent's defensive line to create scoring opportunities. This action, known as a Line Break, is a critical indicator of offensive effectiveness and tactical performance, yet previous studies have mainly focused on shots or goal opportunities rather than on how teams break the defensive line. In this study, we develop a machine learning model to predict Line Breaks using event and tracking data from the 2023 J1 League season. The model incorporates 189 features, including player positions, velocities, and spatial configurations, and employs an XGBoost classifier to estimate the probability of Line Breaks. The proposed model achieved high predictive accuracy, with an AUC of 0.982 and a Brier score of 0.015. Furthermore, SHAP analysis revealed that factors such as offensive player speed, gaps in the defensive line, and offensive players' spatial distributions significantly contribute to the occurrence of Line Breaks. Finally, we found a moderate positive correlation between the predicted probability of being Line-Broken and the number of shots and crosses conceded at the team level. These results suggest that Line Breaks are closely linked to the creation of scoring opportunities and provide a quantitative framework for understanding tactical dynamics in football.
\end{abstract}
\begin{document}

\flushbottom
\maketitle
%
%
\thispagestyle{empty}


\section*{Introduction}

Since the 2000s, data analysis in football has evolved significantly with the introduction of event data. Event data provide detailed records of all on-the-ball actions that occur during a match, such as passes, shots, tackles, and dribbles, along with their timestamps. By utilizing these data, analysts can aggregate key match statistics, such as pass success rates, number of shots, and ball possession percentages, to examine players' and teams' playing styles and tactical tendencies. These statistical summaries, commonly referred to as ``match stats,'' have become essential tools for evaluating player and team performance. However, because event data mainly capture actions involving the ball, they offer limited insight into players' positional relationships, off-the-ball movements, and tactical formations that evolve throughout the game.

With recent advances in tracking technology, it has become possible to capture player and ball movements more objectively through the use of tracking data. Tracking data provide high-resolution records of player and ball positions throughout an entire match, enabling more detailed and efficient analysis of game dynamics than was previously possible \cite{Lucey2014}.
Combining event and tracking data provides deeper tactical and positional insights \cite{Klemp2021, Vidal-Codina2022}.

In particular, automatic and quantitative analytical approaches that incorporate machine learning, statistical modeling, or hybrid combinations of both have attracted increasing attention in recent years. For example, goal prediction models based on the Poisson distribution \cite{poisson}, as well as those utilizing machine learning and deep learning techniques \cite{learning}, have been proposed. Traditionally, most studies have focused on predicting match outcomes such as wins, losses, or goals. However, because football is a low-scoring sport with many draws, predicting match results remains a complex and challenging task. Consequently, recent studies have shifted their focus from overall outcomes to the analysis of individual plays such as passes, shots, and defensive actions \cite{Neves2024, machine}. Building on this trend, metrics such as VAEP (Valuing Actions by Estimating Probabilities) \cite{VAEP}, which quantitatively evaluates player actions based on the probability of scoring or conceding within the next five actions, have been introduced. Furthermore, VDEP (Valuing Defense by Estimating Probabilities) \cite{VDEP} extends this approach by incorporating the positional data of all players and the ball to predict ball recoveries and the likelihood of being attacked within five actions, thereby enabling a quantitative assessment of team defense.

In addition, spatial evaluation methods that focus on player positioning and control of space have gained increasing attention. These approaches quantify each player's influence on the pitch by calculating their area of control based on positional coordinates. For instance, analyses using Voronoi diagrams, minimum time-to-ball models \cite{Rein2016, Taki1996}, or Gaussian distributions \cite{Kijima2014} have enabled the visualization of how much space individual players cover. More recently, studies have examined players' off-ball movements, particularly their ability to create space for teammates, often referred to as space creation ability \cite{Spearman2018, Fernandez2018}. Such spatial approaches provide a powerful means of quantitatively capturing tactical contributions that are not directly reflected in explicit events such as shots or goals.

However, existing models primarily focus on predicting outcomes such as goals scored, goals conceded, or effective attacks, and have not yet addressed the concept of Line Breaks. A Line Break refers to a critical offensive event in football where the attacking team successfully penetrates the opponent's defensive line, thereby creating a potential goal-scoring opportunity. The importance of Line Breaks lies in their direct impact on goal chances when they occur. They often result in situations such as one-on-one opportunities with the goalkeeper or unmarked crossing positions. Conversely, preventing Line Breaks is crucial for the defending side, as the organization and coordination of the defensive line play a decisive role in determining match outcomes. Moreover, the occurrence of Line Breaks is closely related to the spatial structure of the pitch and how available space is utilized. Recent studies in other sports, such as basketball \cite{Kono2024} and ultimate frisbee \cite{Iwashita2024}, have begun integrating quantitative spatial evaluations into their predictive models, suggesting the potential value of applying similar approaches in football.

Therefore, this study aims to construct a machine learning model that predicts the probability of Line Breaks by utilizing event and tracking data from the 2023 J1 League season. The model incorporates a wide range of features, including players' positional coordinates, velocities, and spatial areas quantified using Voronoi diagrams. By utilizing these detailed data, the study seeks to quantitatively identify the factors contributing to the occurrence of Line Breaks and to enhance the understanding of tactical dynamics in football matches. This approach allows for the identification of effective offensive strategies and defensive countermeasures. Furthermore, by examining the relationship between the predicted probability of being Line-Broken and match-related indicators such as the number of shots and crosses conceded, this study explores team-specific tactical tendencies and characteristics.

\section*{Results and Discussion}

\subsection*{Example of Prediction Results}
\begin{figure}[h]
\centering
\includegraphics[width=\linewidth]{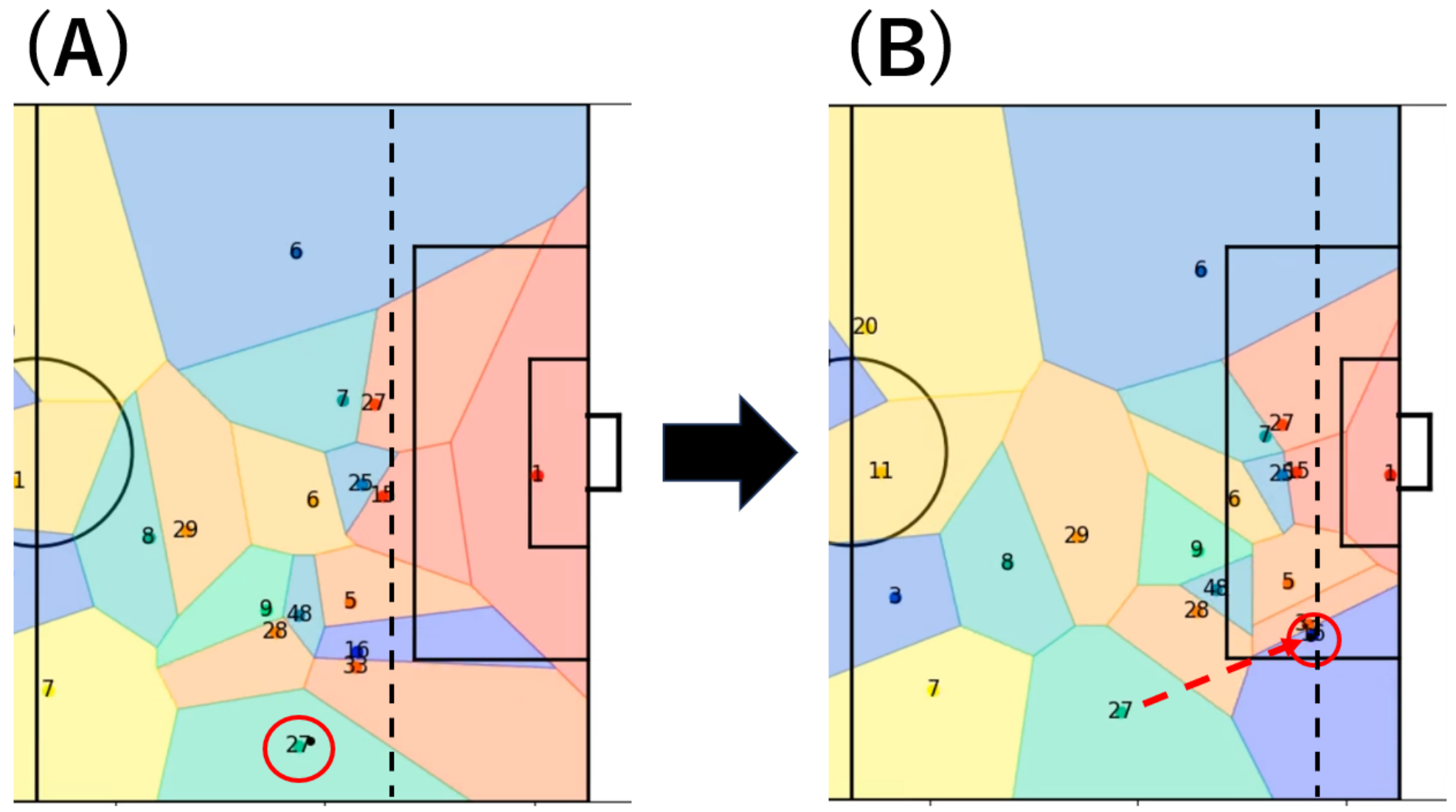}
\caption{Example of a situation in which the proposed model predicted a Line Break with high probability, which was subsequently observed in the match.\\
(A) Moment when Cerezo Osaka's No. 27 (Capixaba) makes a pass.\\
(B) Moment when Cerezo Osaka's No. 16 (Maikuma) receives the pass and immediately delivers a cross.}
\label{fig:example}
\end{figure}

First, we show how our proposed model predicts a Line Break.
Figure \ref{fig:example} illustrates a scene from the 30th round of the J1 League match between Yokohama F. Marinos and Cerezo Osaka. In this situation, Cerezo Osaka launched a crucial attacking play aimed at scoring, capturing the exact moment when a Line Break occurred.  
In Fig.~\ref{fig:example}(A), Cerezo Osaka's No. 27, Capixaba, holds the ball in front of the opposing defensive line and delivers a key pass that initiates the attack. At this moment, the proposed model predicted a high probability (0.870) that this pass would result in a Line Break. This indicates that the model accurately recognized the situation as a critical phase likely to lead to a Line Break. 

Subsequently, Fig.~\ref{fig:example}(B) shows No. 16, Maikuma, making a forward run behind the defensive line and immediately crossing the ball after receiving Capixaba's pass. In this scene, Maikuma successfully breaks through the defensive line and makes contact with the ball in the open space behind the defenders. This situation fully satisfies the definition of a Line Break, confirming that an actual Line Break occurred during this play.

\subsection*{Model Evaluation Results}
Next, we focus on how accurately our model can predict Line Breaks.
The model evaluation results are presented in Table \ref{tab:f1}. From this table, it can be observed that the XGBoost model developed in this study demonstrated solid performance in predicting Line Breaks.

\begin{table}[H]
  \caption{Model Performance Evaluation.}
  \label{tab:f1}
  \centering
  \begin{tabular}{lc}
    \hline
    Evaluation Metric & Value  \\
    \hline \hline
    AUC & 0.982 \\
    Brier score & 0.015 \\
    F1 score & 0.150 \\
    \hline
  \end{tabular}
\end{table}

First, the model achieved a high AUC value of 0.982, indicating excellent discriminative ability in classifying positive and negative Line Break instances. This result suggests that the selection of features and the model design were appropriate, highlighting the potential usefulness of the proposed model in predicting Line Breaks events of significant tactical importance in football.

On the other hand, the Brier score was as low as 0.015, indicating that the predicted probabilities generated by the model showed minimal deviation from the actual outcomes. This finding reveals that the model not only achieved high classification accuracy but also produced highly reliable probabilistic predictions. Such reliability is particularly important when using the predicted probability of Line Breaks to support tactical decision-making during matches.

\begin{figure}[H]
\centering
\includegraphics[width=10cm]{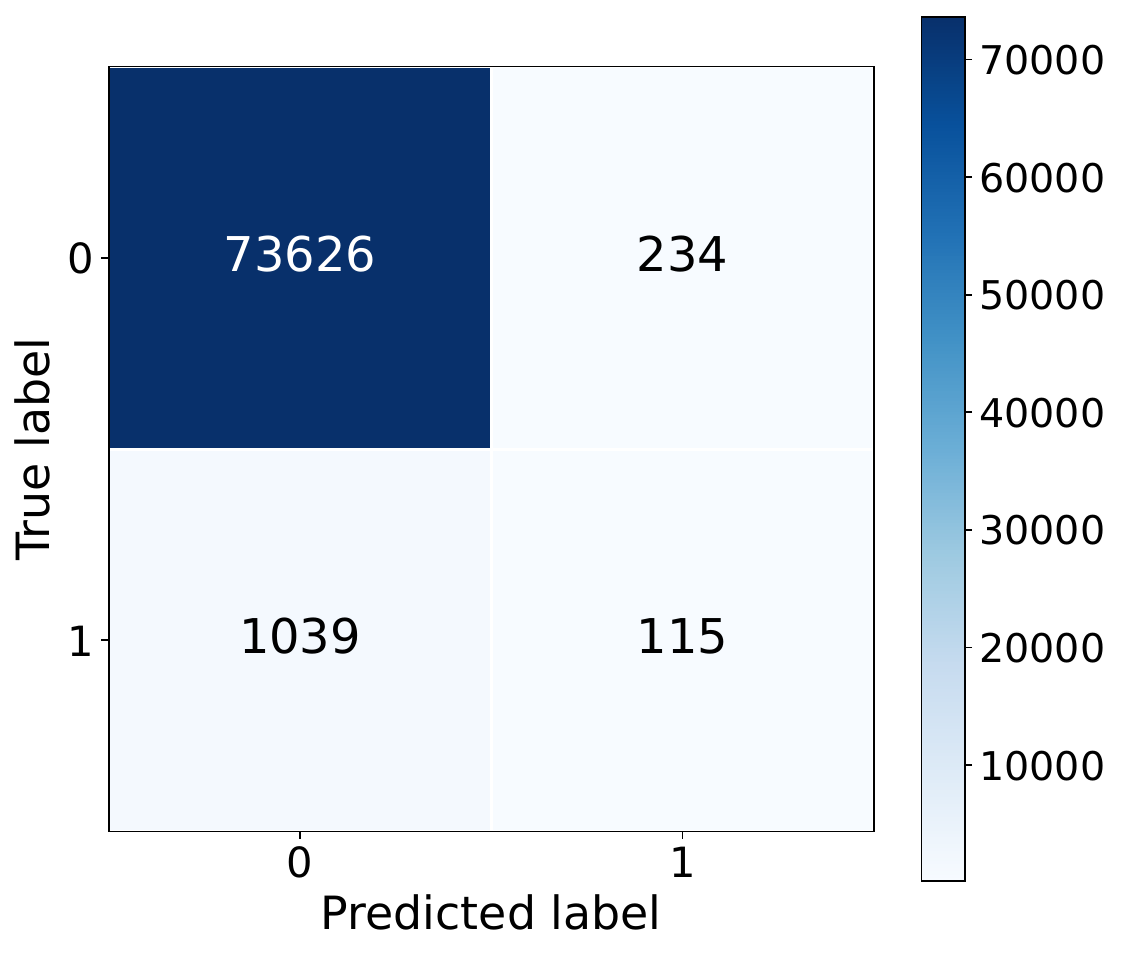}
\caption{Confusion matrix for the Line Break prediction results.}
\label{fig:confusion_matrix}
\end{figure}

However, the F1 score was relatively low at 0.150, indicating that the model tends to overlook positive Line Break instances. This limitation is largely due to the low frequency of Line Break events in the dataset, which causes the model to be biased toward the majority negative class. To provide a clearer picture of this imbalance, Fig. \ref{fig:confusion_matrix} presents the confusion matrix for the Line Break prediction results. Among the cases where the true label was 0 (no Line Break), the model correctly predicted negative in 73,626 instances (True Negatives, TN), while incorrectly predicting positive in 234 cases (False Positives, FP). On the other hand, among the cases where the true label was 1 (Line Break), the model correctly identified 115 instances as positive (True Positives, TP), but misclassified 1,039 cases as negative (False Negatives, FN). These results demonstrate that, although the model effectively suppresses false alarms, detecting positive Line Break events remains challenging due to class imbalance. Moreover, because the occurrence of Line Breaks is influenced by a complex combination of factors, it is possible that the current model cannot fully capture all of these intricate patterns.

Additionally, this study intentionally excluded features related to the receiver of the pass. This design choice was made to account for potential Line Breaks that could occur if the receiving player were different, thereby allowing the model to represent a broader range of in-game scenarios. While this approach enabled more flexible and realistic analysis, it may have also contributed to the reduction in the F1 score.

Overall, based on the AUC and Brier score results, the model demonstrates a reliable level of predictive performance for Line Breaks. Nonetheless, the low F1 score indicates that further improvement is needed to reduce false negatives and enhance the model's ability to detect positive instances.

\subsection*{Feature Importance}
\begin{figure}[H]
\centering
\includegraphics[width=\linewidth]{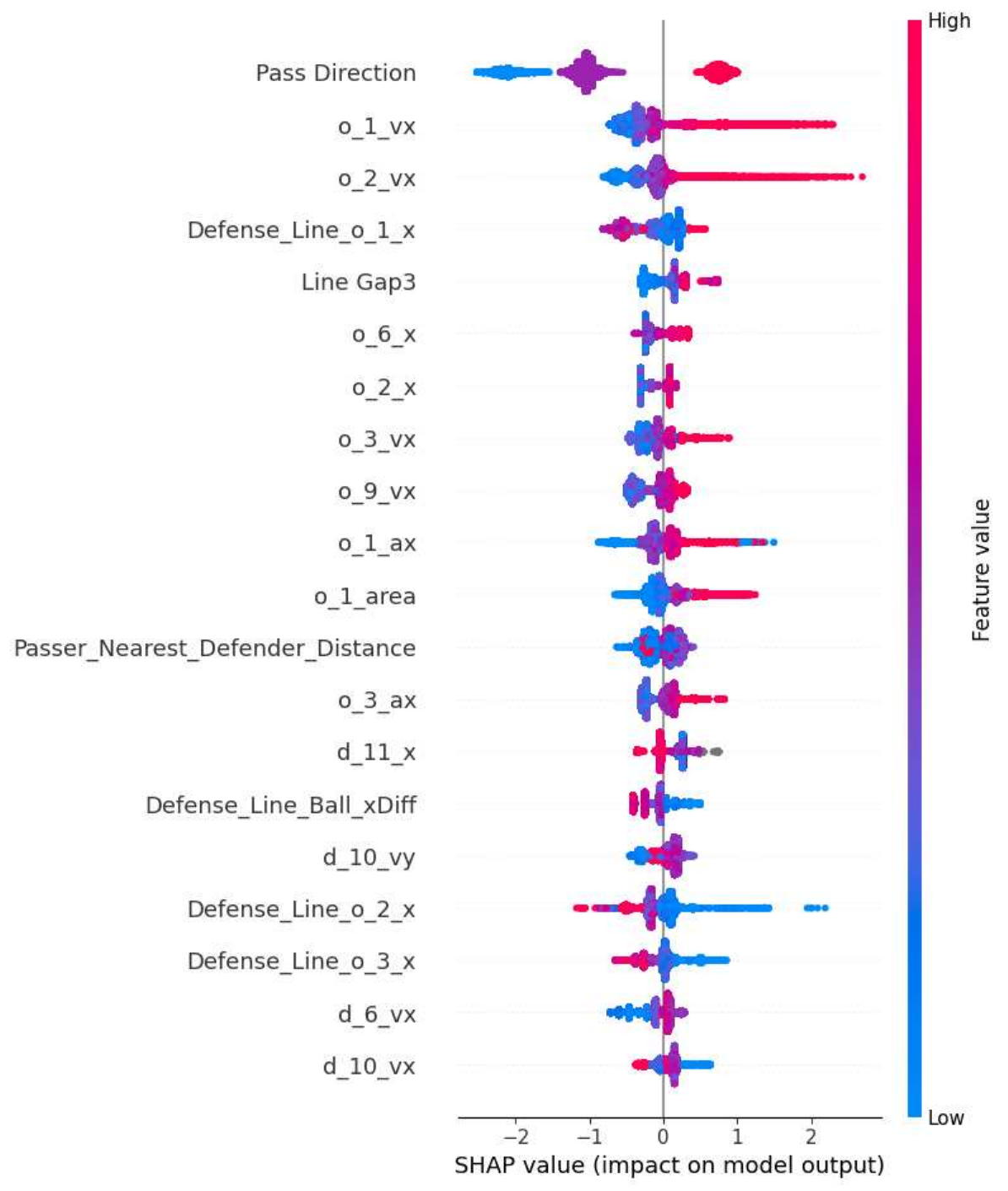}
\caption{Feature importance derived from SHAP analysis.}
\label{fig:shap}
\end{figure}

Our proposed model showed high accuracy.
However, it is also important to understand which features contribute most to the model's predictions.
To address this, we analyzed feature importance using SHAP (SHapley Additive exPlanations) \cite{shap}.
SHAP is a method that applies the concept of Shapley values from game theory to machine learning. In game theory, the Shapley value addresses the problem of how to fairly distribute rewards among players who contribute differently in a cooperative game. In the context of machine learning, this concept is reinterpreted as determining how to allocate the model’s predicted value among features with varying levels of contribution.
Figure \ref{fig:shap} shows which features are most influential in predicting the probability of Line Breaks.   

In Fig.~\ref{fig:shap}, features are listed from top to bottom in descending order of their contribution to the prediction results. A positive SHAP value on the horizontal axis indicates that the corresponding feature contributes to increasing the probability of Line Breaks, whereas a negative SHAP value indicates that it contributes to decreasing it. The color bar represents the magnitude and sign of each feature’s value, where red denotes higher feature values, while blue denotes lower ones. This visualization enables an intuitive understanding of how each feature influences the likelihood of Line Breaks.

\begin{figure}[H]
\centering
\includegraphics[width=\linewidth]{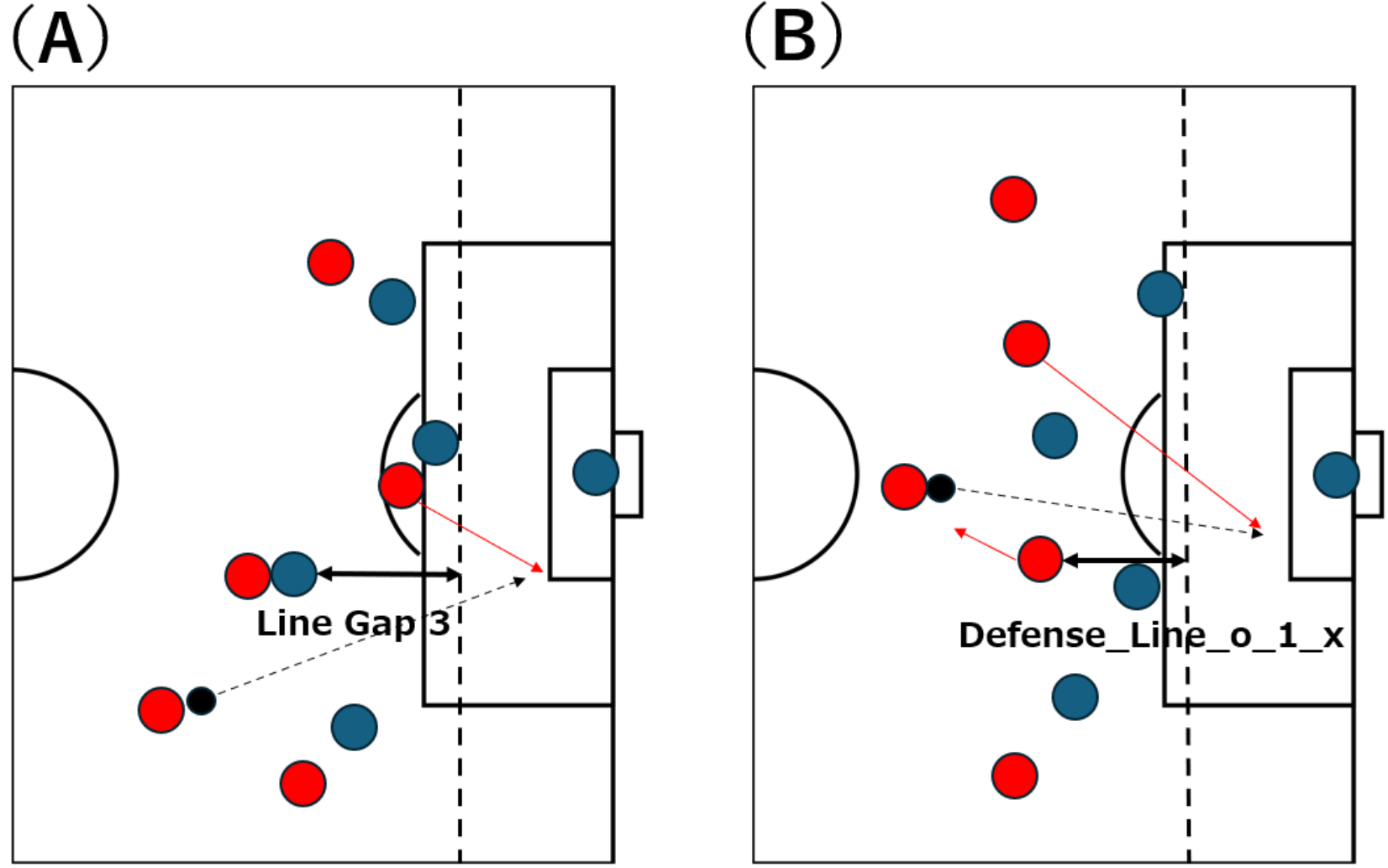}
\caption{(A) Example where a large Line\_Gap3 value leads to the occurrence of a Line Break.\\
(B) Example where a large Defense\_Line\_o\_i\_x value leads to the occurrence of a Line Break.}
\label{fig:gap}
\end{figure}

From the results shown in Fig.~\ref{fig:shap}, several key observations can be made.  
First, the most important feature identified was ``Pass Direction.'' It is natural that forward passes play a crucial role in successfully achieving Line Breaks, and this result supports that intuition.  

Next, features such as ``o\_1\_vx'' and ``o\_2\_vx,'' which represent the velocities of the first and second closest offensive players to the opponent's goal in the x-direction, exhibited high SHAP values, indicating that the movements of forward players significantly contribute to successful Line Breaks. When attacking players actively attempt to penetrate the defensive line, the success rate of Line Breaks upon receiving passes increases. This suggests that the faster the offensive players move toward the opponent's goal, the higher the probability of Line Breaks occurring.  

Features such as ``Line Gap3'' and ``Defense\_Line\_o\_1\_x'' shown in Fig.~\ref{fig:gap} (refer to \textbf{Definition of Line Breaks} for their definitions) indicate that gaps within the defensive line and positional relationships between offensive and defensive players contribute to the occurrence of Line Breaks.  
``Line Gap3'' represents the x-coordinate difference between the rearmost defender (d\_10) and the fourth defender from the back (d\_7). A larger difference indicates a greater likelihood of Line Breaks occurring. Specifically, when defending with a four-back formation, if one defender is pulled out of position, a gap emerges within the defensive line, making it easier for the opponent to break through (Fig.~\ref{fig:gap}(A)).  

Similarly, ``Defense\_Line\_o\_1\_x'' represents the x-coordinate difference between the last defender (d\_10) and the foremost offensive player (o\_1). When this value is small, meaning the offensive player is close to the defensive line, Line Breaks are more likely to occur. However, as indicated by the red points on the right side of the ``Defense\_Line\_o\_1\_x'' plot (Fig.~\ref{fig:shap}), Line Breaks can also occur when this value is large. This corresponds to cases where the foremost forward (o\_1) moves backward, as suggested by the light-blue points on the right side of the ``o\_1\_ax'' plot where o\_1\_ax denotes the acceleration of the foremost attacker in the direction of the opponent’s goal (Fig.~\ref{fig:shap}). In such situations, the movement of o\_1 creates space in the front line, which can then be exploited by the second or third attacking players running behind the defense (Fig.~\ref{fig:gap}(B)).  
These results reveal that both the gaps within the defensive line and the relative positioning between offensive and defensive players have a substantial impact on the risk of Line Breaks.  

Furthermore, ``o\_1\_area'' represents the Voronoi region area of the offensive player. A larger area implies the presence of more open space in front of the forward, suggesting a favorable condition for initiating a Line Break. In practice, when offensive players secure a larger Voronoi region, it indicates effective utilization of the space behind the defense, thereby increasing the likelihood of breaking through the defensive line.  
Thus, the extent to which offensive players can secure advantageous space plays a critical role in determining the occurrence of Line Breaks.

\subsection*{Correlation Between Teams' Defensive Exposure and Predicted Probability of Being Line Broken}

\begin{figure}[htbp]
\centering		
\includegraphics
	[width=\linewidth]{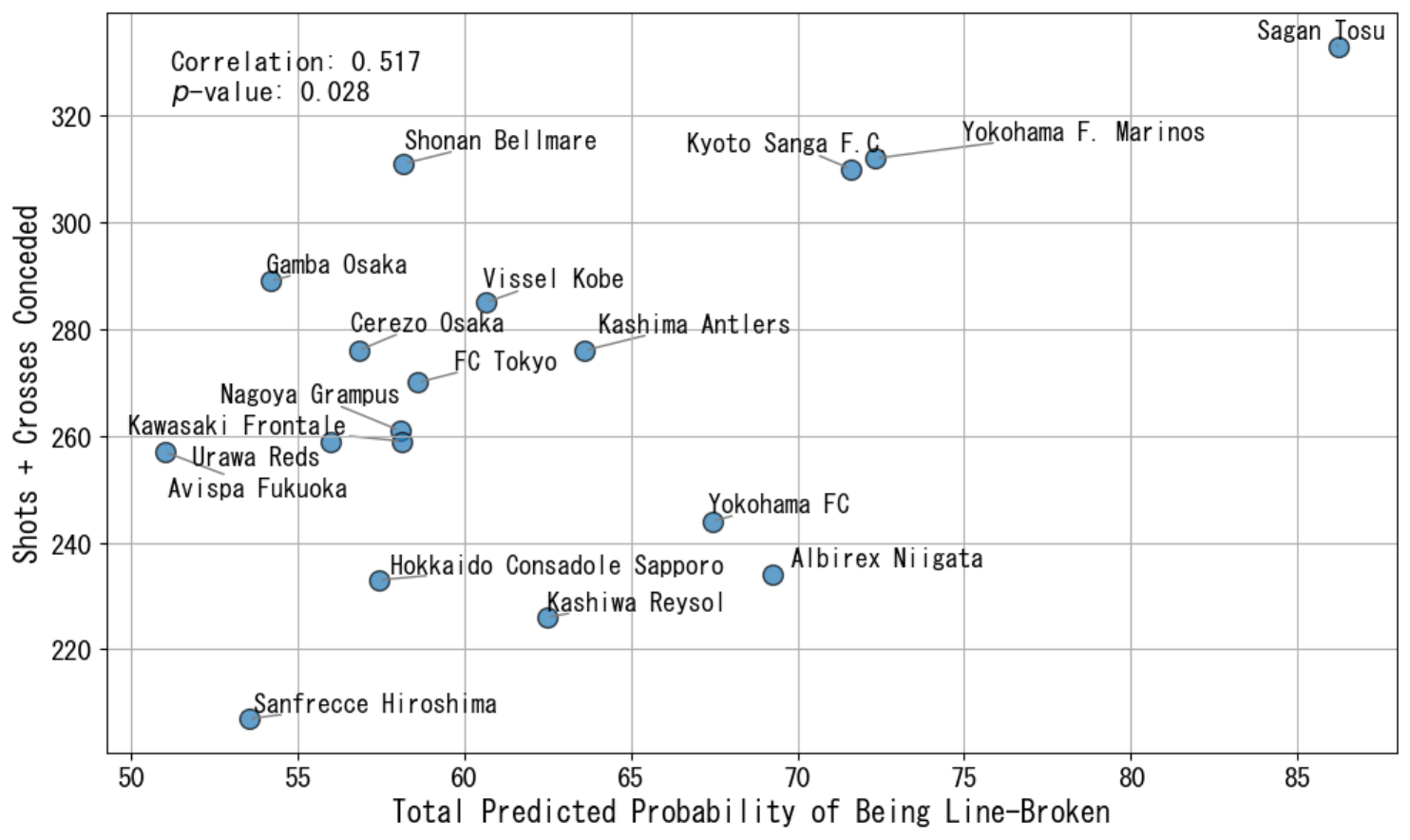}	
\caption{Relationship between the total number of crosses and shots conceded and the predicted probability of being Line-Broken ($r$ = 0.517, $p$ = 0.028).}
\label{fig:shot_cross}
\end{figure}

Finally, we investigate how Line Breaks are related to team performance.
To this end, we analyzed the relationship between each team's total number of crosses and shots conceded and their predicted probability of being Line-Broken (i.e., the probability of conceding a Line Break).
Figure \ref{fig:shot_cross} illustrates this relationship.
The vertical axis represents the total number of crosses and shots conceded over five matches for each team, while the horizontal axis represents the total predicted probability of being Line-Broken. The scatter plot shows a moderate positive correlation between the two indicators ($r$ = 0.517, $p$ = 0.028). This positive relationship can be explained by the fact that when a Line Break occurs, situations such as one-on-one opportunities with the goalkeeper or unmarked crossing chances often arise. Therefore, teams with a higher probability of being Line-Broken tend to allow more crosses and shots from their opponents.

To better understand each team’s pressing tendencies, we referenced Table \ref{tab:press}, provided by Football LAB \cite{FootballLab2023}. This table summarizes five variables, high pressing index, gain rate, defensive success rate, number of shots conceded, and shot success rate, from left to right. It allows for a multifaceted evaluation of each team's pressing tendencies, as well as both the benefits and risks of their high pressing strategies. Teams are listed from top to bottom in descending order of their high pressing index. In this context, ``pressing'' refers to a series of continuous defensive actions in which players apply pressure by rapidly closing down on the opponent in possession of the ball. Furthermore, ``high pressing'' is defined as pressing actions initiated higher up the pitch, specifically above the opponent’s defensive-midfield line.

\begin{table}[H]
\centering
\caption{Summary of High Pressing Strategies (Source: \cite{FootballLab2023}).}
\label{tab:press}
\resizebox{\textwidth}{!}{
\begin{tabular}{lccccc}
\hline
\textbf{Team} & \textbf{High Pressing Index} & \textbf{Gain Rate (\%)} & \textbf{Defensive Success Rate (\%)} & \textbf{Shots Conceded Rate (\%)} & \textbf{Shot Conversion Rate (\%)} \\
\hline \hline
Yokohama F. Marinos & \textbf{68} & 35.2 & 46.2 & 9.2 & 11.7 \\
Sagan Tosu & \textbf{64} & 33.3 & 44.0 & 10.3 & 8.3 \\
Kyoto Sanga F.C. & 63 & 33.7 & 46.9 & 6.5 & 10.4 \\
Sanfrecce Hiroshima & 60 & \textbf{40.0} & \textbf{49.6} & \textbf{5.0} & 21.1 \\
Vissel Kobe & 57 & 37.0 & 47.9 & 6.7 & 9.5 \\
Kawasaki Frontale & 57 & 36.0 & 47.3 & 8.5 & 10.6 \\
Hokkaido Consadole Sapporo & 54 & 39.3 & 48.2 & 6.8 & 19.5 \\
Urawa Reds & 51 & 36.6 & 46.0 & 5.7 & 7.8 \\
Albirex Niigata & 50 & 38.3 & 48.0 & 6.9 & 13.7 \\
FC Tokyo & 49 & 35.4 & 44.9 & 7.7 & 19.8 \\
Shonan Bellmare & 48 & 34.3 & 44.0 & 6.9 & 19.5 \\
Cerezo Osaka & 47 & 32.8 & 43.6 & 7.1 & 13.3 \\
Nagoya Grampus & 46 & 32.7 & 43.1 & 6.4 & 14.1 \\
Avispa Fukuoka & 42 & 33.5 & 40.7 & 6.7 & 15.4 \\
Kashima Antlers & 40 & 34.9 & 44.1 & 6.6 & 18.8 \\
Gamba Osaka & 38 & 35.7 & 45.1 & 8.0 & 18.1 \\
Kashiwa Reysol & 36 & 30.9 & 38.9 & 7.9 & 20.8 \\
Yokohama FC & 31 & 34.4 & 44.8 & 9.5 & 18.9 \\
\hline
\end{tabular}
}
\end{table}

We examined several teams in detail based on Fig.~\ref{fig:shot_cross} and Table \ref{tab:press}. As shown by the high pressing index in Table \ref{tab:press}, Sagan Tosu and Yokohama F. Marinos are characterized by aggressive pressing and fast vertical attacks. While this attacking style enables them to create numerous scoring opportunities during matches, it also tends to stretch their formation, making gaps more likely to appear in the defensive line. These structural weaknesses in defense increase their vulnerability to counterattacks when possession is lost or when the press is bypassed, thereby raising the risk of being Line-Broken. Figure \ref{fig:shot_cross} also confirms that Tosu and Marinos not only concede a high number of crosses and shots but also have relatively high predicted probabilities of being Line-Broken. This suggests that the trade-off between offensive aggressiveness and defensive vulnerability influences the occurrence of Line Breaks.

In contrast, Sanfrecce Hiroshima records the fewest crosses and shots conceded among all teams and also exhibits a low predicted probability of being Line-Broken. Although Hiroshima, like Marinos, employs a high pressing style and can sometimes be exposed to counterattacks, the strong defensive capabilities of their three-back system, comprising Sasaki, Araki, and Shiotani, combined with the high gain rate and defensive success rate shown in Table \ref{tab:press}, allow them to maintain a stable defense even against counterattacks. Furthermore, other teams that employ a three-back system, such as Hokkaido Consadole Sapporo, Gamba Osaka, Nagoya Grampus, and Avispa Fukuoka, also tend to show lower probabilities of being Line-Broken. This is likely because teams with a three-back formation can form a five-back defensive block when pushed deep into their own half, thereby reducing the space available for opponents to execute Line Breaks.

In conclusion, this study developed a machine learning model to predict Line Breaks in football and analyzed the key factors influencing their occurrence using feature importance analysis.  
The results revealed that multiple elements, such as player speed, available space, distances from defenders, and gaps within the defensive line, play significant roles in the occurrence of Line Breaks.  
Our model achieved high predictive accuracy, with an AUC of 0.982 and a Brier score of 0.015, demonstrating strong capability in estimating Line Break probabilities.  
However, the relatively low F1 score (0.150) highlights the inherent difficulty of predicting rare Line Break events.  
One contributing factor to this result is the intentional exclusion of features related to the pass receiver, a design choice made to consider potential Line Breaks involving different receivers and to capture broader in-game scenarios.

Moreover, the analysis revealed a moderate positive correlation between the predicted probability of being Line-Broken and the total number of crosses and shots conceded, highlighting team-specific tactical tendencies.  
For instance, offensively oriented teams such as Sagan Tosu and Yokohama F. Marinos demonstrated higher Line Break risks due to aggressive pressing and fast vertical attacks, which tend to create gaps within the defensive line.  
Conversely, defensively well-structured teams, particularly those employing a three-back system such as Sanfrecce Hiroshima, Gamba Osaka, and Nagoya Grampus, showed reduced Line Break susceptibility.  
These findings suggest that our proposed model provides valuable insights into team tactics, spatial organization, and defensive compactness.

To mitigate Line Break risks, teams may lower the defensive line to minimize space behind them, while raising the line can increase compactness and limit the opponent's freedom.  
In future work, we aim to extend the model to incorporate additional offensive and defensive events, ultimately identifying optimal defensive line positioning under different match situations and contributing to more flexible tactical decision-making in football analytics.

\section*{Methods}
\subsection*{Definition of Line Breaks}\label{subsec:def_lb}

\begin{figure}[htbp]
\centering		
\includegraphics
	[width=\linewidth]{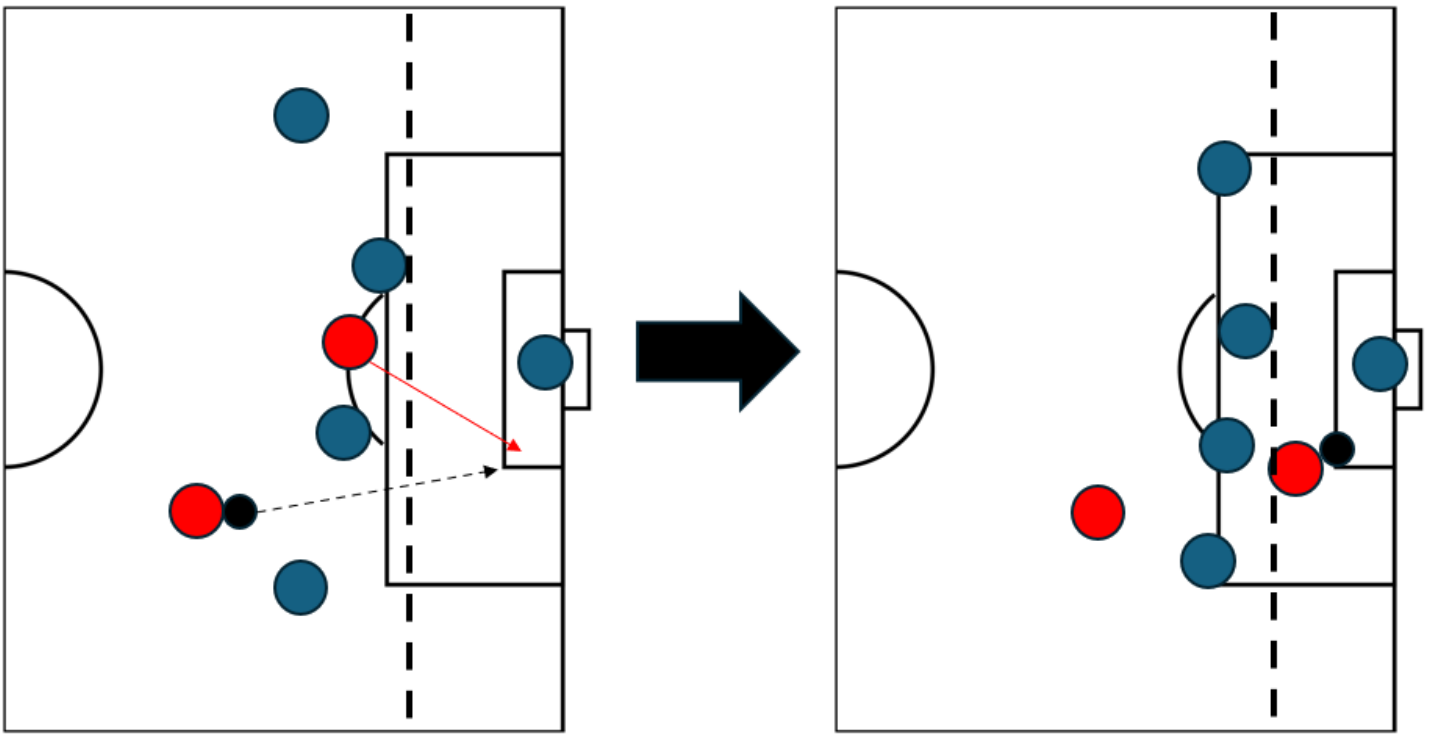}	
\caption{Definition of a Line Break (black: ball, red: offensive team, blue: defensive team).}
\label{fig:break}
\end{figure}

In this study, a Line Break is defined as follows (as illustrated in Fig.~\ref{fig:break}): at the moment the pass is made, both the ball and the receiving player are positioned in front of the offside line (referred to in this study as the defensive line), but when the receiver makes contact with the ball, they have moved beyond, or broken through, the defensive line. The defensive line is defined as an imaginary line (the black dashed line in Fig.~\ref{fig:break}) drawn based on the position of the second-last defensive player relative to the goal line (the boundary line adjacent to the goal).

\subsection*{Data Used}
In this study, event data, tracking data, and match information data from 90 matches (Rounds 25–34) of the 2023 Meiji Yasuda J1 League season were provided by Data Stadium Inc., Japan and used for analysis. DataStadium has a contract with J1 League to collect and sell data. We use the dataset for this research under the permission by DataStadium. The event data include action labels such as passes and shots recorded at 1/30-second intervals, along with the corresponding ball position coordinates. The tracking data consist of the positional coordinates of all players and the ball recorded at 25 Hz, and the match information data contain results such as goals scored and conceded. Although teams usually switch their attacking directions between the first and second halves, all coordinate data were standardized so that the attacking direction of the team in possession (the offensive team) always corresponded to the positive x-axis direction (left to right), ensuring analytical consistency. Data for which the ball's position coordinates were missing, making it difficult to identify the receiver, were excluded from the event data in advance.

For predicting Line Breaks with the machine learning model developed in this study, four rounds (36 matches) were used as training data and one round (9 matches) as test data, following a fivefold cross-validation procedure to generate predictions for all 45 matches.

Furthermore, among all successful passes, the model predicts whether a Line Break occurs for each pass classified as a ``pass action.'' In this study, pass actions include ``home pass,'' ``away pass,'' ``through pass,''and ``flick-on'' as defined in the provided dataset. A ``home pass'' and an ``away pass'' refer to passes made by players on the home and away teams, respectively. A ``through pass'' refers to a pass played into open space in front of the receiving player rather than directly to their feet. A ``flick-on'' refers to a pass in which a player lightly redirects a teammate's pass, altering its trajectory to continue the play.

\subsection*{Overview of Model Implementation}
\subsubsection*{Feature Engineering}

The features used for model training are listed in Table \ref{tab:features}. The feature set consists of a total of 189 dimensions, including all players' positional coordinates, velocities, and information related to the defensive line. In Table \ref{tab:features}, the x-axis is parallel to the touchline, and the y-axis is parallel to the goal line. The prefix ``o'' denotes the offensive team in possession of the ball, while ``d'' denotes the defending team without the ball. For example, features labeled as o\_1 or d\_3 indicate player indices assigned in order of proximity to the opponent’s goal within each team. In other words, players with smaller numbers typically represent forwards (FW), whereas those with larger numbers generally correspond to goalkeepers (GK) or defenders (DF). In all cases, the defending team's goal is standardized to be located in the positive x-axis direction.

\begin{table}[htbp]
  \caption{List of Features Used in the Model.}
  \label{tab:features}
  \centering
  \resizebox{\textwidth}{!}{  
  \begin{tabular}{l p{9cm} c}
    \hline
    \textbf{Name} & \textbf{Description} & \textbf{Dimensions} \\
    \hline \hline
    Start Ball X & x-coordinate of the passer (ball position) & 1 \\
    Start Ball Y & y-coordinate of the passer (ball position) & 1 \\
    Through Pass & Whether the pass is a through pass & 1 \\
    Flick On & Whether the pass is a flick-on & 1 \\
    Direct Pass & Whether the pass is a one-touch (direct) pass & 1 \\
    Pass Direction & Direction of the pass (forward, lateral, backward) & 1 \\
    Passer\_Nearest\_Defender\_Distance & Distance between the passer and the nearest defender & 1 \\
    Line Gap1–3 & x-coordinate differences between the reference defender on the defensive line and other defenders (Fig.~\ref{fig:line}(A)) & 3 \\
    Defense\_Line\_Ball\_xDiff & Difference in x-coordinate between the defensive line and the ball for the defensive team & 1 \\
    Offense\_Line\_Ball\_xDiff & Difference in x-coordinate between the offensive line and the ball for the offensive team & 1 \\
    Defense\_Lines\_xDiff & Difference in x-coordinate between both teams’ defensive lines & 1 \\
    o\_i\_x, d\_i\_x & x-coordinates of all players & 22 \\
    o\_i\_y, d\_i\_y & y-coordinates of all players & 22 \\
    o\_i\_vx, d\_i\_vx & Players’ velocities in the x-direction & 22 \\
    o\_i\_vy, d\_i\_vy & Players’ velocities in the y-direction & 22 \\
    o\_i\_ax, d\_i\_ax & Players’ accelerations in the x-direction & 22 \\
    o\_i\_ay, d\_i\_ay & Players’ accelerations in the y-direction & 22 \\
    o\_i\_area, d\_i\_area & Areas occupied by each player derived from Voronoi diagrams & 22 \\
    Defense\_Line\_o\_i\_x & Difference in x-coordinate between the reference defender on the defensive line and each offensive player (Fig.~\ref{fig:line}(B)) & 11 \\
    Defense\_Line\_o\_i\_y & Difference in y-coordinate between the reference defender on the defensive line and each offensive player (Fig.~\ref{fig:line}(B)) & 11 \\
    \hline
    Total &  & 189 \\
    \hline
  \end{tabular}
  }
\end{table}

\begin{figure}[htbp]
\centering		
\includegraphics
	[width=\linewidth]{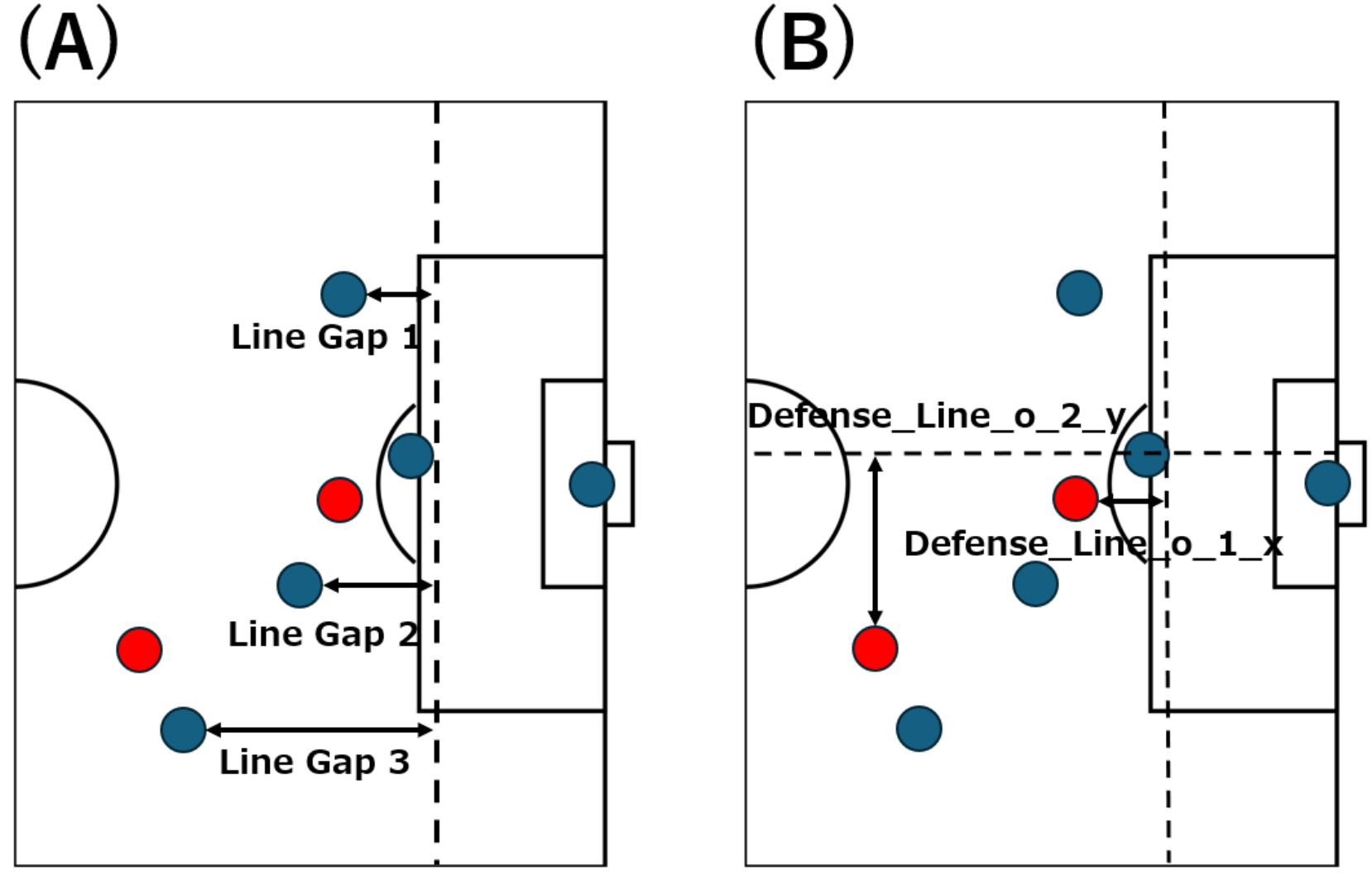}	
\caption{Description of Features.\\(A) Line Gap.\\(B) Defense\_Line\_o\_i\_x, Defense\_Line\_o\_i\_y.}
\label{fig:line}
\end{figure}

In this study, the Voronoi diagram was computed based on the players' positional coordinates using the SciPy library in Python.  
A Voronoi diagram is a method for dividing a plane into regions based on multiple points (called *sites* or *generating points*) distributed on the plane.  
Each region is characterized by the property that any point within it is closer to its corresponding generating point than to any other.  

Specifically, given a finite subset \( P \subset X \) in a metric space \( (X, d) \), for each generating point \( p \in P \), the set of points in \( X \) that are closest to \( p \) among all points in \( P \) is defined as:  

\begin{equation}
V(p) = \{ x \in X \mid \forall q \in P \, [\, d(x, p) \leq d(x, q) \,] \}
\end{equation}

This set \( V(p) \) is called the (Voronoi) region of \( p \), and the collection of all such regions for every \( p \in P \) constitutes the Voronoi diagram of \( P \).  
Here, \( d(x, p) \) represents the Euclidean distance between a point \( x \) and a generating point \( p \).  
In practice, the Voronoi diagram can be constructed by drawing the boundary lines (perpendicular bisectors) that connect points equidistant from neighboring generating points, thereby partitioning the plane into regions.

\begin{figure}[h]
\centering		
\includegraphics
	[width=\linewidth]{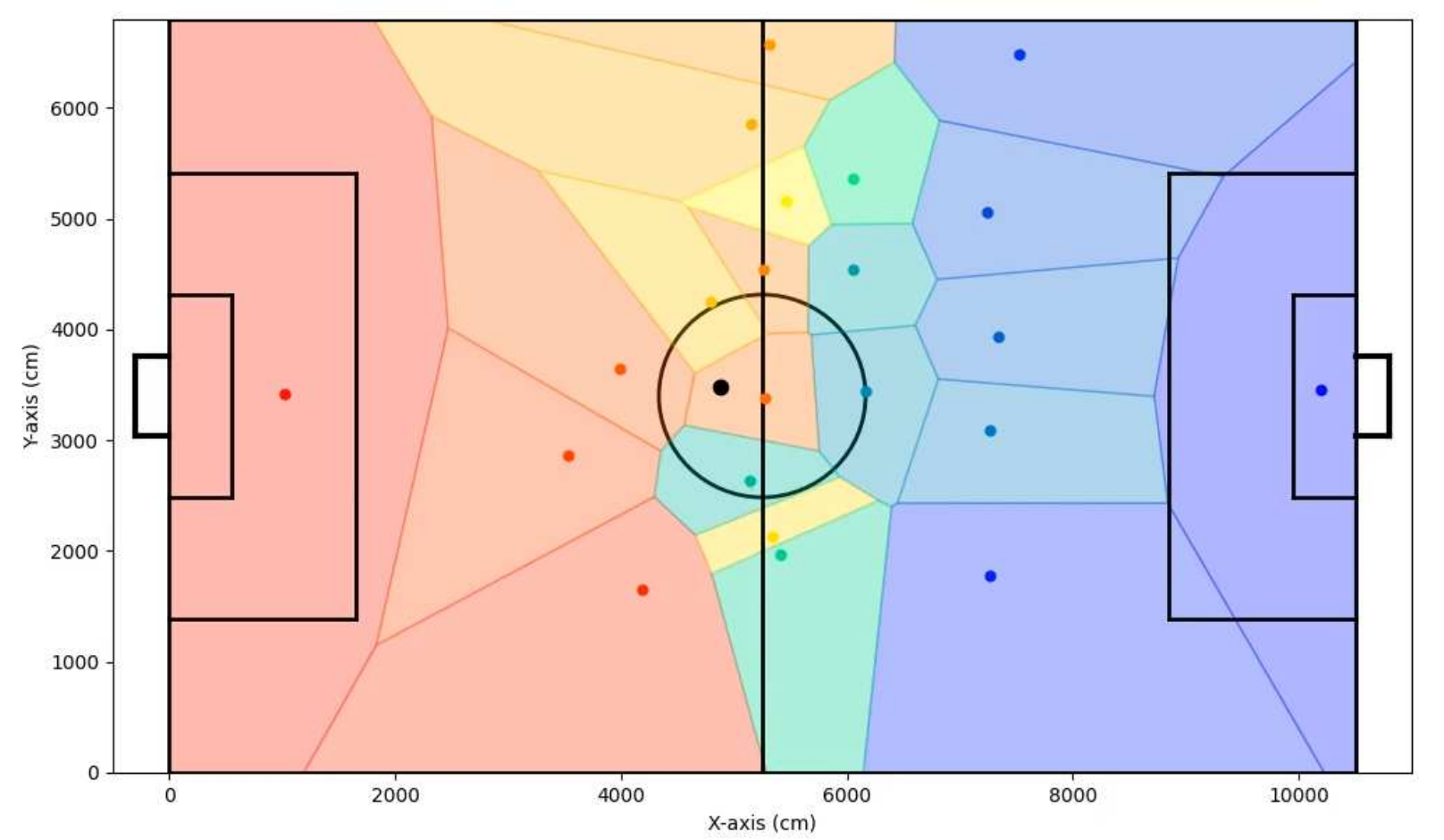}	
\caption{Visualization of Each Player’s Space Using Voronoi Diagram.}
\label{fig:voronoi}
\end{figure}

Figure \ref{fig:voronoi} shows an example of a Voronoi diagram generated from actual player position data. Each player is represented as a point (generating point). Since every region in the diagram satisfies the condition that any point within the region is closer to its own generating point than to any other generating point in different regions, each region can be interpreted as the space controlled by the corresponding player (the generating point) within that area.

\subsubsection*{Model Training and Evaluation}

For the classification of Line Break occurrences, we employed XGBoost (eXtreme Gradient Boosting) \cite{xgboost}, which has been used in previous studies \cite{VAEP, VDEP}.  
XGBoost is a learning algorithm that efficiently and scalably optimizes gradient boosting using decision trees.  
The primary reason for selecting this algorithm in the present study is that XGBoost provides a highly optimized implementation of the gradient boosting technique, excelling in both computational efficiency and predictive accuracy.  
Specifically, XGBoost trains an ensemble of decision trees sequentially, improving the model at each step by minimizing prediction errors.

The learning process of the XGBoost algorithm can be summarized as follows:  

\begin{enumerate}
    \item First, the true labels \( y_i \) and features \( f_i \) are defined.
    \item The initial prediction value \( \hat{y}_i \) is set, typically initialized to 0.5.
    \item For each training sample, the error \( f_i(x_i) \) from the true label \( y_i \) is calculated, and the product of this error and the learning rate \( \mu \) is used to update the leaf weights (i.e., the amount of model improvement). This process is repeated through conditional branching based on feature values within the decision tree structure.
    \item The leaf weights of each decision tree are adjusted in the direction that minimizes the overall model error. Through these iterations, the model's predictive performance is progressively enhanced.
    \item The final prediction value \( \hat{y}_i \) is obtained by adding the sum of the weighted outputs from all trees to the initial prediction:
\end{enumerate}

\begin{equation}
\hat{y}_i = 0.5 \ (\text{initial value}) + \sum_{k=1}^{K} \mu f_k(x_i)
\end{equation}

where \( K \) denotes the number of boosting iterations.

Through this continuous boosting process, the model's generalization performance improves, enabling the accurate prediction of complex events such as Line Breaks.  
Moreover, XGBoost incorporates several optimization techniques, including parallel computation, early stopping, and tree depth regularization, that allow for scalable model construction even with large datasets.  
In football analytics, XGBoost has also been widely applied in prior studies, such as in the goal and concession prediction models of VAEP \cite{VAEP} and the defensive effectiveness and ball recovery prediction models of VDEP \cite{VDEP}.

In the dataset used for this study, a total of 38,970 passes across all teams were analyzed, of which 584 were identified as positive instances of Line Breaks.  
Since the model outputs continuous probability values ranging from 0 to 1, a classification threshold of 0.5 was adopted: predictions with probabilities of 0.5 or higher were labeled as positive (Line Breaks), while those below 0.5 were labeled as negative.

To evaluate the performance of the model, three metrics were used: AUC (Area Under the Curve), Brier score, and F1 score.  
AUC represents the area under the ROC (Receiver Operating Characteristic) curve and measures the model’s ability to distinguish between positive and negative instances.  
A value close to 1 indicates strong discriminative performance, whereas a value near 0.5 implies random prediction.  
The ROC curve itself visually expresses the trade-off between the true positive rate (sensitivity) and the false positive rate (1 – specificity), and it is widely used to intuitively assess classifier performance.

The Brier score evaluates the accuracy of probabilistic predictions by calculating the mean squared difference between the predicted probabilities and the actual outcomes.  
This metric quantifies how well the model predicts probabilities, where lower scores (closer to 0) indicate higher predictive accuracy.

Finally, the F1 score, defined as the harmonic mean of precision and recall, was used to comprehensively assess the model’s ability to predict the positive class.  
The F1 score is particularly important for imbalanced datasets, as it considers the trade-off between false positives and false negatives.  
This metric ranges from 0 to 1, with values closer to 1 indicating superior predictive performance for positive instances.

Using these evaluation metrics, the performance of the proposed XGBoost model was examined from multiple perspectives to determine how accurately it could predict the occurrence of Line Breaks.

\subsection*{Data Availability}
The J-League event and player-tracking dataset used in this study is not publicly available due to contractual restrictions. It may be obtained from the corresponding author (G. Ichinose) upon reasonable request and with approval from DataStadium Inc., Japan.

\subsection*{Code availability}
The underlying code for this study is not publicly available but may be made available to qualified researchers on reasonable request from the corresponding author.

\bibliography{citations-Football}

\begin{thebibliography}{10}
\urlstyle{rm}
\expandafter\ifx\csname url\endcsname\relax
  \def\url#1{\texttt{#1}}\fi
\expandafter\ifx\csname urlprefix\endcsname\relax\def\urlprefix{URL }\fi
\expandafter\ifx\csname doiprefix\endcsname\relax\def\doiprefix{DOI: }\fi
\providecommand{\bibinfo}[2]{#2}
\providecommand{\eprint}[2][]{\url{#2}}

\bibitem{Lucey2014}
\bibinfo{author}{Lucey, P.}, \bibinfo{author}{Bialkowski, A.}, \bibinfo{author}{Monfort, M.}, \bibinfo{author}{Carr, P.} \& \bibinfo{author}{Matthews, I.}
\newblock \bibinfo{title}{Quality vs quantity: Improved shot prediction in soccer using strategic features from spatiotemporal data}.
\newblock In \emph{\bibinfo{booktitle}{Proceedings of the MIT Sloan Sports Analytics Conference}} (\bibinfo{year}{2014}).

\bibitem{Klemp2021}
\bibinfo{author}{Klemp, M.}, \bibinfo{author}{Wunderlich, F.} \& \bibinfo{author}{Memmert, D.}
\newblock \bibinfo{journal}{\bibinfo{title}{In-play forecasting in football using event and positional data}}.
\newblock {\emph{\JournalTitle{Scientific Reports}}} \textbf{\bibinfo{volume}{11}}, \bibinfo{pages}{24139}, \doiprefix\url{10.1038/s41598-021-03157-3} (\bibinfo{year}{2021}).

\bibitem{Vidal-Codina2022}
\bibinfo{author}{Vidal-Codina, F.}, \bibinfo{author}{Evans, N.}, \bibinfo{author}{El~Fakir, B.} \& \bibinfo{author}{Billingham, J.}
\newblock \bibinfo{journal}{\bibinfo{title}{Automatic event detection in football using tracking data}}.
\newblock {\emph{\JournalTitle{Sports Engineering}}} \textbf{\bibinfo{volume}{25}}, \bibinfo{pages}{18}, \doiprefix\url{10.1007/s12283-022-00381-6} (\bibinfo{year}{2022}).

\bibitem{poisson}
\bibinfo{author}{Karlis, D.} \& \bibinfo{author}{Ntzoufras, I.}
\newblock \bibinfo{journal}{\bibinfo{title}{Analysis of sports data by using bivariate poisson models}}.
\newblock {\emph{\JournalTitle{Journal of the Royal Statistical Society: Series D (The Statistician)}}} \textbf{\bibinfo{volume}{52}}, \bibinfo{pages}{381--393}, \doiprefix\url{https://rss.onlinelibrary.wiley.com/doi/10.1111/1467-9884.00366} (\bibinfo{year}{2003}).

\bibitem{learning}
\bibinfo{author}{R.~Bunker, C.~Y.} \& \bibinfo{author}{Fujii, K.}
\newblock \bibinfo{journal}{\bibinfo{title}{Machine learning for soccer match result prediction}}.
\newblock {\emph{\JournalTitle{arXiv preprint arXiv:2403.07669}}} \bibinfo{pages}{1–43}, \doiprefix\url{https://arxiv.org/abs/2403.07669} (\bibinfo{year}{2024}).

\bibitem{Neves2024}
\bibinfo{author}{Neves, T.}, \bibinfo{author}{Rodrigues, F.}, \bibinfo{author}{Saleiro, P.} \& \bibinfo{author}{Gama, J.}
\newblock \bibinfo{title}{Towards a foundation large events model for soccer}, \doiprefix\url{10.1007/s10994-024-06606-y} (\bibinfo{year}{2024}).

\bibitem{machine}
\bibinfo{author}{Fujii, K.}
\newblock \emph{\bibinfo{title}{Machine Learning in Sports: Open Approach for Next Play Analytics}} (\bibinfo{publisher}{Springer}, \bibinfo{year}{2025}).

\bibitem{VAEP}
\bibinfo{author}{T.~Decroos, J. V.~H., L.~Bransen} \& \bibinfo{author}{Davis, J.}
\newblock \bibinfo{journal}{\bibinfo{title}{Actions speak louder than goals: Valuing player actions in soccer}}.
\newblock {\emph{\JournalTitle{KDD}}} \bibinfo{pages}{1851--1861} (\bibinfo{year}{2019}).

\bibitem{VDEP}
\bibinfo{author}{K.~Toda, K.~K., M.~Teranishi} \& \bibinfo{author}{Fujii, K.}
\newblock \bibinfo{journal}{\bibinfo{title}{Evaluation of soccer team defense based on prediction models of ball recovery and being attacked}}.
\newblock {\emph{\JournalTitle{PLOS ONE 17(1): e0263051}}}  (\bibinfo{year}{2022}).

\bibitem{Rein2016}
\bibinfo{author}{Rein, R.}, \bibinfo{author}{Raabe, D.}, \bibinfo{author}{Perl, J.} \& \bibinfo{author}{Memmert, D.}
\newblock \bibinfo{title}{Evaluation of changes in space control due to passing behavior in elite soccer using voronoi‑cells}.
\newblock In \emph{\bibinfo{booktitle}{Proceedings of the 10th International Symposium on Computer Science in Sports (ISCSS)}}, vol. \bibinfo{volume}{392} of \emph{\bibinfo{series}{Advances in Intelligent Systems and Computing}}, \bibinfo{pages}{179--183}, \doiprefix\url{10.1007/978-3-319-24560-7_23} (\bibinfo{publisher}{Springer}, \bibinfo{year}{2016}).

\bibitem{Taki1996}
\bibinfo{author}{Taki, T.}, \bibinfo{author}{Hasegawa, J.} \& \bibinfo{author}{Fukumura, T.}
\newblock \bibinfo{title}{Development of motion analysis system for quantitative evaluation of teamwork in soccer games}.
\newblock In \emph{\bibinfo{booktitle}{Proc.\ 3rd IEEE International Conference on Image Processing (ICIP‑96)}}, vol.~\bibinfo{volume}{3}, \bibinfo{pages}{815--818} (\bibinfo{year}{1996}).

\bibitem{Kijima2014}
\bibinfo{author}{Kijima, A.}, \bibinfo{author}{Yokoyama, K.}, \bibinfo{author}{Shima, H.} \& \bibinfo{author}{Yamamoto, Y.}
\newblock \bibinfo{journal}{\bibinfo{title}{Emergence of self-similarity in football dynamics}}.
\newblock {\emph{\JournalTitle{The European Physical Journal B}}} \textbf{\bibinfo{volume}{87}}, \bibinfo{pages}{41}, \doiprefix\url{10.1140/epjb/e2014-40987-5} (\bibinfo{year}{2014}).

\bibitem{Spearman2018}
\bibinfo{author}{Spearman, W.}
\newblock \bibinfo{title}{Beyond expected goals}.
\newblock In \emph{\bibinfo{booktitle}{Proceedings of the 12th MIT Sloan Sports Analytics Conference}}, \bibinfo{pages}{1--17} (\bibinfo{year}{2018}).

\bibitem{Fernandez2018}
\bibinfo{author}{Fernandez, J.} \& \bibinfo{author}{Bornn, L.}
\newblock \bibinfo{title}{Wide open spaces: A statistical technique for measuring space creation in professional soccer}.
\newblock In \emph{\bibinfo{booktitle}{Proceedings of the 12th MIT Sloan Sports Analytics Conference}} (\bibinfo{year}{2018}).

\bibitem{Kono2024}
\bibinfo{author}{Kono, T.} \& \bibinfo{author}{Fujii, K.}
\newblock \bibinfo{title}{Mathematical models for off-ball scoring prediction in basketball}.
\newblock In \emph{\bibinfo{booktitle}{Proceedings of the International Workshop on Machine Learning and Data Mining for Sports Analytics}}, \doiprefix\url{10.1007/978-3-031-86692-0_4} (\bibinfo{publisher}{Springer}, \bibinfo{year}{2024}).

\bibitem{Iwashita2024}
\bibinfo{author}{Iwashita, S.}, \bibinfo{author}{Scott, A.}, \bibinfo{author}{Umemoto, R.}, \bibinfo{author}{Ding, N.} \& \bibinfo{author}{Fujii, K.}
\newblock \bibinfo{title}{Space evaluation based on pitch control using drone video in ultimate}.
\newblock \bibinfo{howpublished}{arXiv preprint arXiv:2409.14588} (\bibinfo{year}{2024}).

\bibitem{shap}
\bibinfo{author}{Lundberg, S.~M.} \& \bibinfo{author}{Lee, S.-I.}
\newblock \bibinfo{journal}{\bibinfo{title}{A unified approach to interpreting model predic tions}}.
\newblock {\emph{\JournalTitle{NIPS'17: Proceedings of the 31st International Conference on Neural Information Processing Systems}}} \bibinfo{pages}{4768--4777} (\bibinfo{year}{2017}).

\bibitem{FootballLab2023}
\bibinfo{author}{{Football LAB}}.
\newblock \bibinfo{howpublished}{\url{https://www.football-lab.jp/summary/team_style/j1?year=2023&data=66}} (\bibinfo{year}{2023}).
\newblock \bibinfo{note}{Accessed: 2025-07-28}.

\bibitem{xgboost}
\bibinfo{author}{Chen, T.} \& \bibinfo{author}{Guestrin, C.}
\newblock \bibinfo{journal}{\bibinfo{title}{Xgboost: A scalable tree boosting system}}.
\newblock {} \bibinfo{pages}{785--794} (\bibinfo{year}{2016}).

\end{thebibliography}

\section*{Acknowledgements (not compulsory)}

This study utilized data provided by Data Stadium Inc.. 
The authors express their gratitude to Data Stadium Inc. for providing access to the event and tracking data of the J1 League 2023 season.  
The data used in this research are subject to copyright and licensing agreements.  
Appropriate copyright notices are displayed in accordance with the provider’s requirements:  
``Data provided by Data Stadium Inc.'' and ``© J STATS''.

\section*{Author contributions statement}

G.I. negotiated with Data Stadium Inc. for data use in this study.
S.Y. conceived the idea, developed the model, and analyzed the results.
G.I. and J.I. determined the overall direction of the study, and S.Y. conducted additional analyses.
S.Y. drafted the initial manuscript, and G.I. and J.I. revised and refined it.
All authors reviewed and approved the final version of the manuscript.

\section*{Additional information}
\textbf{Competing interests} The authors declare no competing interests. 
The corresponding author is responsible for submitting a competing interests statement.

\end{document}